\documentclass{article}

\usepackage{color}
\usepackage[dvipsnames]{xcolor}
\usepackage{colortbl}
\usepackage{graphicx}
\usepackage{booktabs}
\usepackage{xspace}
\usepackage[utf8]{inputenc} % allow utf-8 input
\usepackage[T1]{fontenc}    % use 8-bit T1 fonts
\usepackage[colorlinks,
            linkcolor=VioletRed,      
            anchorcolor=VioletRed, 
            citecolor=VioletRed,       
            ]{hyperref}

\usepackage{multirow}
\usepackage{natbib}  % DO NOT CHANGE THIS AND DO NOT ADD ANY OPTIONS TO IT
\usepackage{wrapfig}
\usepackage{subfigure}

\definecolor{Gray}{gray}{0.85}

\title{\Large  MME-RealWorld: Could Your Multimodal LLM \\ Challenge High-Resolution Real-World Scenarios \\ that are Difficult for Humans?}

\usepackage[preprint]{neurips_2025}

\usepackage{hyperref}       % hyperlinks
\usepackage{url}            % simple URL typesetting
\usepackage{booktabs}       % professional-quality tables
\usepackage{amsfonts}       % blackboard math symbols
\usepackage{nicefrac}       % compact symbols for 1/2, etc.
\usepackage{microtype}      % microtypography
\usepackage{xcolor}         % colors

\usepackage{geometry} % For better page layout
\usepackage{booktabs} % For professional table rules (\toprule, \midrule, \bottomrule)
\usepackage{tabularx} % For tables with text wrapping columns
\usepackage{amsmath} % For math symbols like \le
\usepackage{enumitem} % For customizing lists if needed (optional here, but good practice)
\usepackage{array}
\newcolumntype{Y}{>{\centering\arraybackslash}X}

\usepackage{cleveref}

\title{Zoom-Refine: Boosting High-Resolution \\ Multimodal Understanding via \\ Localized Zoom and Self-Refinement}

\author{Xuan Yu$^1$  \quad 
Dayan Guan$^1$\thanks{Corresponding Author.}  \quad
Yanfeng Gu$^1$ \\
$^1$ Harbin Institute of Technology
}

\begin{document}

\maketitle

\begin{abstract}
Multimodal Large Language Models (MLLM) often struggle to interpret high-resolution images accurately, where fine-grained details are crucial for complex visual understanding. We introduce Zoom-Refine, a novel training-free method that enhances MLLM capabilities to address this issue. Zoom-Refine operates through a synergistic process of \textit{Localized Zoom} and \textit{Self-Refinement}. In the \textit{Localized Zoom} step, Zoom-Refine leverages the MLLM to provide a preliminary response to an input query and identifies the most task-relevant image region by predicting its bounding box coordinates. During the \textit{Self-Refinement} step, Zoom-Refine then integrates fine-grained details from the high-resolution crop (identified by \textit{Localized Zoom}) with its initial reasoning to re-evaluate and refine its preliminary response. Our method harnesses the MLLM's inherent capabilities for spatial localization, contextual reasoning and comparative analysis without requiring additional training or external experts. Comprehensive experiments demonstrate the efficacy of Zoom-Refine on two challenging high-resolution multimodal benchmarks.
Code is available at \href{https://github.com/xavier-yu114/Zoom-Refine}{\color{magenta}github.com/xavier-yu114/Zoom-Refine}
\end{abstract}

\section{Introduction}
\label{sec:introduction}

The advent of Multimodal Large Language Models (MLLMs) marks a significant milestone in artificial intelligence, demonstrating remarkable understanding and reasoning capabilities by combining textual and visual information~\cite{gpt4v_openai2023, team2023gemini}. These models have immense potential across diverse applications, from autonomous driving and medical image analysis to robotics and human-computer interaction. Significant efforts focus on expanding their performance boundaries through scaling models, data, and training strategies~\cite{chen2024expanding,zhu2025internvl3}. However, a critical bottleneck persists: effectively processing high-resolution images. Real-world visual data is often rich in detail, and the ability to discern fine-grained features \textit{(e.g.}, small objects, subtle textures, intricate text) is essential for accurate comprehension and complex reasoning tasks~\cite{mme_realworld_iclr2025,wang2025divide}. 
Most current MLLMs often struggle in this regard, typically relying on downsampling and cropping high-resolution images to fit fixed-resolution vision encoders, leading to inevitable information loss.

Intriguingly, humans interpret complex and high-resolution visual scenes with remarkable efficiency, suggesting alternative processing strategies. Human visual perception is not a passive and uniform intake of information. Instead, it is an active and dynamic process involving selective attention and refinement. Foundational research on visual cognition elucidates how humans utilize saccadic eye movements to rapidly shift their gaze and direct high-acuity foveal vision (a biological mechanism for "\textit{Localized Zoom}") onto specific regions of interest identified through peripheral processing or task demands~\cite{posner1980orienting, henderson2003human}. This allows for a detailed analysis of critical areas without the need to process the entire scene at maximum resolution simultaneously. Furthermore, human visual understanding often involves a process of hypothesis testing and verification (a perceptual form of "\textit{Self-Refinement}"). Initial interpretations based on a glance or global context are frequently checked and updated based on detailed information gathered from these foveated fixations, allowing ambiguities to be resolved and errors corrected~\cite{findlay2003active, hayhoe2005eye}. This interplay between broad contextual awareness and targeted scrutiny is fundamental to robust human visual intelligence.

\begin{figure*}[t]
% \centering
% \footnotesize
\centering\includegraphics[width=.99\linewidth]{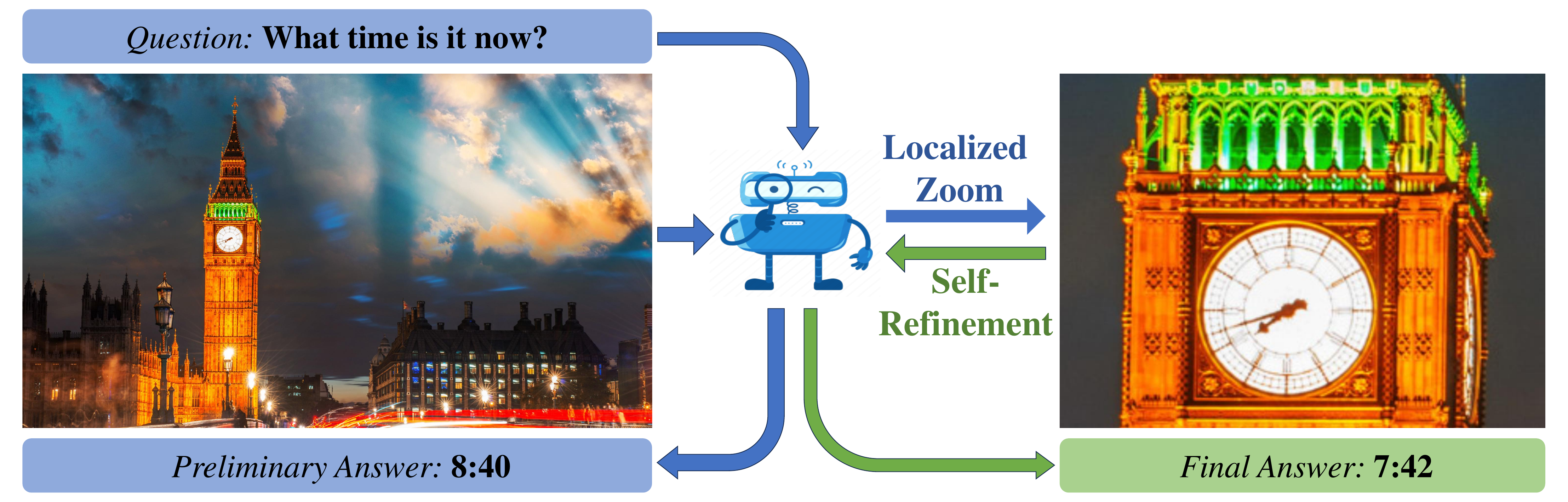}
\vspace{-10pt}
\caption{The framework of Zoom-Refine. The MLLM initially processes an input \textit{Question} and a downsampled image, yielding a \textit{Preliminary Answer}. The \textit{Localized Zoom} step then identifies the most relevant image region, from which a high-resolution crop is extracted. In the subsequent \textit{Self-Refinement} step, this high-resolution crop provides fine-grained visual context to the MLLM, which integrates it with the initial global context to produce a more accurate \textit{Final Answer}.}
\vspace{-5pt}
\label{fig_framework}
\end{figure*}

Inspired by human visual intelligence, we introduce Zoom-Refine, a novel and training-free framework enhancing MLLM capabilities with high-resolution images, as shown in Figure~\ref{fig_framework}. Zoom-Refine operates through a synergistic process of \textit{Localized Zoom} and \textit{Self-Refinement}. Initially, the MLLM receives an input query, which comprises a downsampled version of the original high-resolution image and an associated textual question. During the \textit{Localized Zoom} step, the MLLM provides a preliminary answer and predicts bounding box coordinates for the image region most relevant to the task. A high-resolution crop, corresponding to these predicted bounding-box coordinates, is then extracted from the original high-resolution source image. Subsequently, in the \textit{Self-Refinement} step, this high-resolution crop is presented back to the same MLLM, serving as the fine-grained visual context. The MLLM then integrates this fine-grained information with the global context retained from the initial view, allowing it to re-evaluate its initial hypothesis, identify potential omissions or misinterpretations, and produce a refined, more accurate final output.

It is worth noting that Zoom-Refine requires no additional training or external experts, intrinsically leveraging the MLLM's inherent capabilities for prompt-based reasoning, spatial localization (via bounding box prediction), and contextual integration. We demonstrate the effectiveness of Zoom-Refine through comprehensive experiments on two challenging high-resolution benchmarks: MME-RealWorld~\cite{mme_realworld_iclr2025} and HR-Bench~\cite{wang2025divide}. Our results show significant performance improvements across a variety of demanding multimodal tasks, validating the efficacy of mimicking human-like localized zoom and self-refinement strategies for boosting high-resolution multimodal understanding. 

We summarize the contributions of this work in three aspects. 
\textit{Firstly}, we introduce a novel paradigm for high-resolution multimodal understanding, inspired by human visual intelligence. This paradigm advocates a shift from static holistic processing to a dynamic strategy integrating \textit{Localized Zoom} for fine-grained analysis and \textit{Self-Refinement} for evidence-based correction.
\textit{Secondly}, we instantiate this paradigm with Zoom-Refine, a novel and training-free framework. Zoom-Refine uniquely empowers an MLLM to intrinsically perform task-relevant region localization and subsequent self-correction using high-resolution visual crops, without requiring additional training or external experts.
\textit{Thirdly}, we demonstrate significant performance gains with Zoom-Refine on two challenging high-resolution benchmarks, validating the effectiveness of our method.

\section{Related Works}
\label{sec:related_works}
\textbf{Multimodal Large Language Models (MLLMs).} The field of MLLMs is advancing at an accelerated pace, producing models with increasingly sophisticated vision-language integration. Cutting-edge proprietary systems like GPT-4V~\cite{gpt4v_openai2023} and Gemini~\cite{team2023gemini} exemplify the peak capabilities currently available. Simultaneously, the open-source community continues to push boundaries with state-of-the-art models~\cite{zhu2025internvl3,chen2024expanding,bai2025qwen2,li2024llava}. Beyond these prominent examples, a diverse and rapidly growing ecosystem of MLLMs has emerged, encompassing various large-scale foundational models~\cite{zhang2023internlm, lu2024deepseek, bai2023qwen,liu2023visual,wu2024visionllm}, iterative improvements on established architectures~\cite{xie2024graph,jiang2024maven,li2024cumo,laurenccon2024matters,lin2024vila,lu2025task}, efficient compact designs~\cite{chen2023minigpt, hu2024minicpm,liu2024visual,hu2024matryoshka,li2023otter,tong2024cambrian}, and specialized models targeting challenges like text-rich document understanding or enhanced high-resolution processing~\cite{liu2024textmonkey,chen2024sharegpt4v, li2024hrvqa,wang2024leveraging,zhang2024wings,huang2024aggregate}. Many recent contributions demonstrate improved handling of higher input resolutions compared to earlier generations.  However, despite these advances, efficiently and effectively understanding images at extremely high resolutions (e.g., 4K, 8K, or even larger) remains a significant challenge. 

\textbf{High-Resolution Multimodal Understanding.} Addressing MLLM challenges with high-resolution inputs involves diverse strategies beyond simply scaling input size. Training-based approaches include architectural innovations like InternLM-XComposer2-4KHD~\cite{dong2024internlm}, which uses dynamic patches for up to 4K HD inputs, DualFocus~\cite{cao2024dualfocus}, which trains models for macro-micro integration using specialized data, Visual cot~\cite{shao2024visual}, which trains models for such focused understanding by providing a dataset with intermediate visual reasoning to integrate global and detailed local views and frameworks like SEAL~\cite{wu2024v}, which feature LLM-guided visual search algorithms and underlying visual search models that may require training for specific subtasks. Alternatively, various training-free methods enhance existing MLLMs. These often involve guided exploration and multi-stage processing, such as tree-based search (ZoomEye~\cite{shen2024zoomeye}), MLLM-derived text guidance over image divisions ($\mathrm{DC}^2$~\cite{wang2025divide}), Chain-of-Thought for text-rich images (TextCoT~\cite{luan2024textcot}). In addition, DyFo~\cite{li2025dyfo} and Zoomer~\cite{qian2025zoomer} leverage external visual expert modules for tasks like region selection or feature extraction . While these methods focus on improved region selection or guided information extraction, Zoom-Refine distinctively incorporates an explicit self-refinement step, leveraging fine-grained details from the MLLM-localized zoom to refine and enhance the initial understanding.

\textbf{Self-Refinement.} The concept of enabling models to autonomously review and improve their outputs, known as self-refinement or self-correction~\cite{madaan2023self, pan-etal-2024-automatically}, is crucial for enhancing LLM reliability. Prior approaches achieve correction using external environmental feedback, such as explicit success/failure signals~\cite{shinn2023reflexion,renze2024self}. Latest researches investigate self-correction without external environmental feedback, noting the inherent difficulty LLMs face in identifying their own errors~\cite{li-etal-2024-hindsight, huang2024large, tyen-etal-2024-llms, kamoi-etal-2024-llms}. Methodologies diverge based on the intervention stage: some focus on integrating self-correction capabilities during model training through techniques like reinforcement learning~\cite{kumar2025training,han2024small, zhang2024learning}, while others, more aligned with our approach, design inference-time strategies. These inference-time methods typically employ structured prompting or multi-step pipelines to encourage models to reflect upon, critique, and revise their initial outputs~\cite{dhuliawala2024chain, wu2024large, madaan2023self, kim2023language}. Zoom-Refine contributes to the line of inference-time self-refinement by uniquely grounding the correction process in new, task-relevant visual evidence acquired through its MLLM-guided localized zoom.

\section{Method}
\label{sec:method}

Addressing the limitations of Multimodal Large Language Models (MLLMs) in processing high-resolution images necessitates moving beyond simple downsampling techniques that discard crucial fine-grained details. Inspired by the nature of human visual perception involving selective attention (foveation) and hypothesis refinement, we propose \textbf{Zoom-Refine}, a training-free framework designed to enhance high-resolution multimodal understanding. Zoom-Refine operates in two synergistic stages: \textit{Localized Zoom}, where the MLLM identifies and requests a high-resolution view of a task-relevant image region, and \textit{Self-Refinement}, where the MLLM critically re-evaluates its initial assessment by integrating the detailed local information with the original global context. This approach leverages the inherent capabilities of the MLLM for spatial localization and contextual reasoning without requiring architectural changes or additional supervised training.

\subsection{Problem Definition}
Let $I \in \mathbb{R}^{H \times W \times C}$ be a high-resolution image, where $H$, $W$, and $C$ denote the height, width, and number of channels, respectively, and let $Q$ be a textual question pertaining to image $I$. The core task in high-resolution multimodal understanding is to generate an accurate answer $A$ based on the joint input $(I, Q)$. Standard MLLMs, denoted by $\mathcal{M}$, often utilize vision encoders accepting fixed-size inputs. This necessitates downsampling high-resolution images $I$ to $I_\text{ds}$, where $I_\text{ds} \in \mathbb{R}^{h \times w \times C}$ with $h < H$ and $w < W$. The conventional inference process is typically represented as:
\begin{equation}
    A_{base} = \mathcal{M}(I_\text{ds}, Q)
    \label{eq:baseline}
\end{equation}
This downsampling operation, however, leads to an unavoidable loss of fine-grained visual information present in the original image $I$. Such loss can significantly impair the model's performance, particularly when the correct answer hinges on discerning small objects, interpreting fine text, or recognizing subtle visual patterns. Our objective is thus to develop a method enabling $\mathcal{M}$ to effectively utilize the rich detail within the original high-resolution image $I$, thereby producing an improved final answer $A_\text{final}$ relative to the baseline $A_{base}$, all without necessitating any retraining of the MLLM.

\begin{figure*}[!t]
% \centering
% \footnotesize
\centering\includegraphics[width=1.0\linewidth]{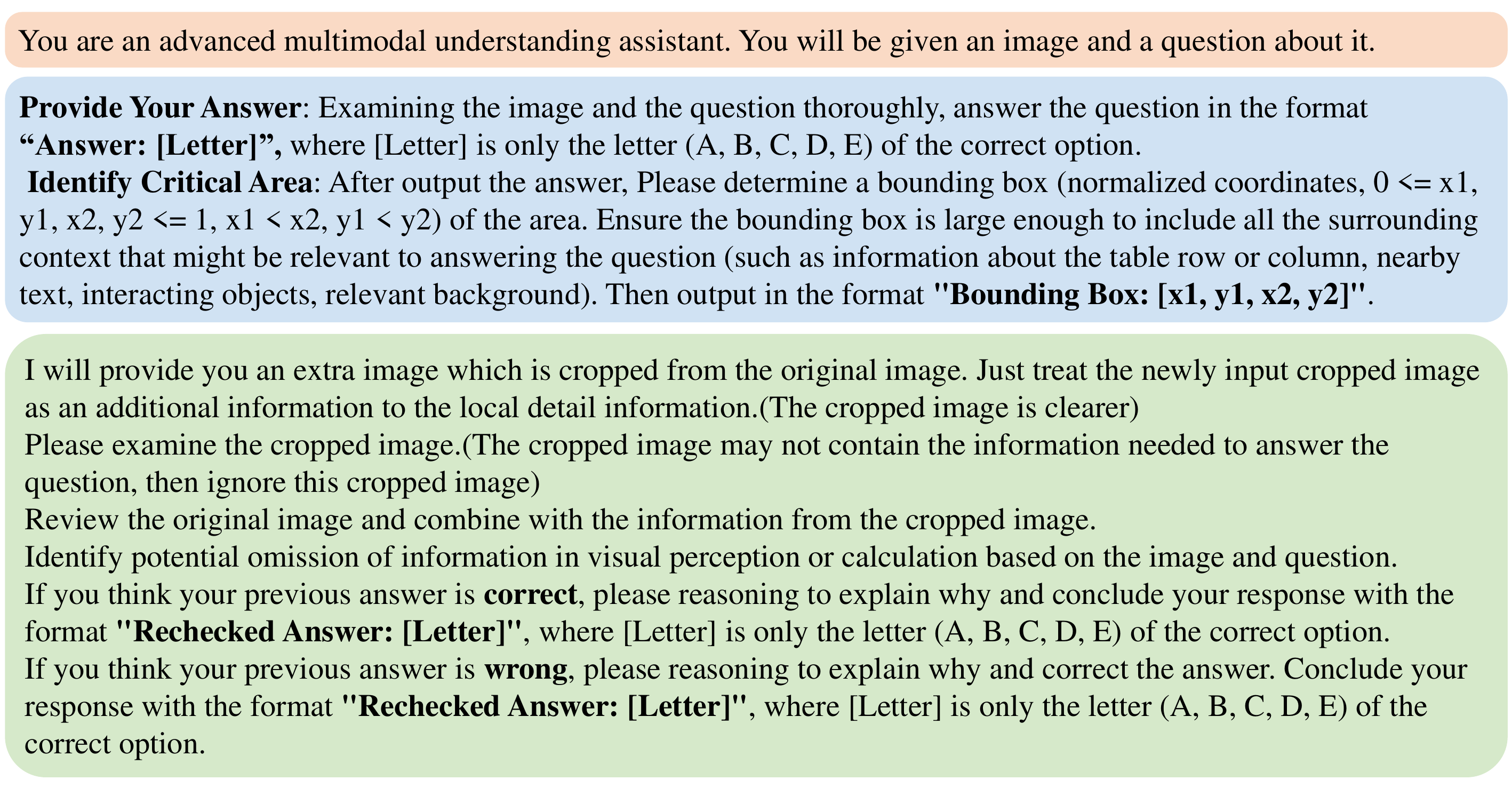}
\vspace{-15pt}
\caption{Prompt engineering of Zoom-Refine. System prompt, "Localized Zoom" prompt, and "Self-refinement" prompt are highlighted in \textcolor[rgb]{0.93, 0.49, 0.19}{orange}, \textcolor[rgb]{0.36, 0.61, 0.84}{blue}, and \textcolor[rgb]{0.44, 0.68, 0.28}{green}, respectively. 
% Best viewed in color.
}
% \vspace{-10pt}
\label{fig_prompt}
\end{figure*}

\subsection{Zoom-Refine: Localized Zoom and Self-Refinement}
\label{sec:zoom_refine_details}

Zoom-Refine enhances high-resolution understanding through a sequential two-stage process, both executed by the same underlying MLLM $\mathcal{M}$. The first stage focuses on generating an initial assessment and identifying a critical region for closer examination. The MLLM $\mathcal{M}$ initially processes the potentially downsampled image $I_\text{ds}$ alongside the question $Q$ to produce a preliminary answer, $A_\text{init}$. Concurrently, or immediately thereafter, the model is prompted via $P_\text{loc}$ to identify the image region most crucial for answering $Q$. This involves predicting a bounding box $B = [x_1, y_1, x_2, y_2]$, specified using normalized coordinates, which delineates this vital area. The prompt guides the model to select a region sufficiently large to include relevant contextual elements. This initial analysis step, leveraging the MLLM's spatial reasoning, can be summarized as:
\begin{equation}
    (A_\text{init}, B) = \mathcal{M}(I_\text{ds}, Q, P_\text{loc})
    \label{eq:stage1}
\end{equation}

Following the prediction of the bounding box $B$, the Localized Zoom is executed by extracting a high-resolution crop, $I_\text{crop}$, directly from the original, full-resolution image $I$ based on the coordinates in $B$. This operation is denoted as:
\begin{equation}
    I_\text{crop} = \mathcal{F}_\text{crop}(I, B)
    \label{eq:crop}
\end{equation}
This extracted crop $I_\text{crop}$ preserves the fine-grained details that might have been lost during the initial downsampling to $I_\text{ds}$.

The second stage involves the Self-Refinement process, where the high-resolution crop $I_\text{crop}$ is presented back to the MLLM as supplementary evidence to facilitate a re-evaluation of its initial hypothesis. The MLLM $\mathcal{M}$ is provided with the history $H_\text{init}$ from the first stage (including $I_\text{ds}$, $Q$, and potentially $A_\text{init}$ along with its reasoning) and the new high-resolution crop $I_\text{crop}$. A carefully designed prompt, $P_\text{refine}$, then directs the MLLM to scrutinize $I_\text{crop}$ for detailed information potentially missed during the analysis of $I_\text{ds}$. The model is explicitly asked to compare the fine-grained details observed in $I_\text{crop}$ with the broader context from $I_\text{ds}$ and its initial reasoning process. Based on this comparative analysis, the MLLM must either reaffirm its initial answer $A_\text{init}$, providing justification, or revise its answer if contradictions or new insights emerge, explaining the rationale for the correction. This guided re-evaluation encourages the model to perform active self-correction. The generation of the final, refined answer $A_\text{final}$ is formalized as:
\begin{equation}
    A_\text{final} = \mathcal{M}(H_\text{init}, I_\text{crop}, P_\text{refine})
    \label{eq:stage2
}
\end{equation}
Thus, the Zoom-Refine framework transforms the standard single-pass inference into a more sophisticated, two-step reasoning procedure, $\text{Zoom-Refine}(I, Q, \mathcal{M})$, adeptly utilizing the MLLM's own capabilities to dynamically focus on pertinent high-resolution details and refine its overall understanding.

\section{Experiments}
\label{experiment}
In this section, we conduct extensive experiments to evaluate the effectiveness of our proposed {Zoom-Refine} method. We primarily show results by applying Zoom-Refine to strong baselines such as InternVL3-78B~\cite{zhu2025internvl3} and InternVL2.5-78B~\cite{chen2024expanding}.
A key aspect of our evaluation methodology is the consistent application of a single, carefully designed prompt for all experiments across all benchmarks and model variations. This prompt, detailed in Figure~\ref{fig_prompt}, is designed to effectively guide the MLLMs without introducing additional task-specific hyperparameters, ensuring a fair and robust assessment of Zoom-Refine's capabilities.

\renewcommand\arraystretch{0.93}
\begin{table}[ht]
\caption{{Experimental results on the reasoning tasks of the MME-Realworld benchmark.} OCR, DT, MO, and AD each indicate a specific subtask: Optical Character Recognition in the Wild, Diagram and Table, Monitoring, and  Autonomous Driving, respectively. Avg and Avg-C indicate the weighted and unweighted average accuracy across subtasks in each subtask. 
}\label{tab:mmerw_reason}
\small
\centering
\begin{tabularx}{1.0\textwidth}{lXXXXXX}
\toprule 
\textbf{Method} & \textbf{OCR} & \textbf{DT} & \textbf{MO} & \textbf{AD} & \textbf{Avg} & \textbf{Avg-C} \\ 
\midrule
% \multicolumn{2}{c}{\textbf{\# QA pairs}} & 500 & 500 & 498 & 1344 & 2842 & 2842 \\ \midrule 
\textbf{InternVL3-78B + Ours} & \textbf{75.40} & \textbf{71.80} & \textbf{47.39} & \textbf{40.22} &\textbf{53.22} & \textbf{58.70} \\
InternVL3-78B~\cite{zhu2025internvl3} & 72.80 & 62.20 & 43.78 & 38.47 & 49.62 & 54.31 \\
 \midrule 
{InternVL2.5-78B + Ours} & {74.20} & {60.80} & {44.38} & {38.99} & {49.97} & {54.59} \\
InternVL2.5-78B~\cite{chen2024expanding} & 71.20 & 53.20 & 37.35 & 36.75 & 45.60 & 49.52  \\
\midrule
Qwen2-VL-7B~\cite{wang2024qwen2} & 63.40 & 48.60 & 33.13 & 31.47 & 40.39 & 44.15 \\
InternVL2-7B~\cite{chen2024internvl} & 57.40 & 39.00 & 43.57 & 29.84 & 38.74 & 42.45 \\ 
GPT-4o~\cite{hurst2024gpt} & 61.40 & 44.80 & 36.51 & 26.41 & 37.61 & 42.28 \\
CogVLM2-8B~\cite{hong2024cogvlm2} & 54.00 & 32.80 & 41.16 & 31.18 & 37.25 & 39.79 \\
InternVL-20B~\cite{chen2024internvl} & 56.80 & 35.40 & 37.35 & 28.94 & 36.48 & 39.62 \\
Cambrian-1-8B~\cite{tong2024cambrian} & 53.20 & 27.40 & 42.37 & 30.73 & 36.16 & 38.43 \\
SliME-8B~\cite{zhang2024beyond} & 53.20 & 29.40 & 36.14 & 31.55 & 35.80 & 37.57 \\
MiniCPM-V2.5-8B~\cite{hu2024minicpm} & 44.00 & 31.80 & 36.95 & 31.03 & 34.50 & 35.95 \\
SliME-13B~\cite{zhang2024beyond}  & 41.00 & 39.00 & 33.13 & 30.80 & 34.46 & 35.98 \\
InternLM-XComposer2.5-7B~\cite{zhang2023internlm} & 53.40 & 41.00 & 17.67 & 29.99 & 33.90 & 35.52 \\ 
GPT-4o-mini~\cite{hurst2024gpt} & 47.00 & 39.80 & 25.81 & 26.79 & 32.48 & 34.85 \\
YI-VL-34B~\cite{young2024yi} & 42.40 & 26.00 & 31.33 & 31.55 & 32.45 & 32.82 \\
LLaVA-Next-8B~\cite{li2024llava} & 55.20 & 23.40 & 21.08 & 30.73 & 32.06 & 32.60 \\
Mini-Gemini-34B-HD~\cite{li2024mini} & 59.20 & 39.20 & 20.48 & 22.84 & 31.73 & 35.43 \\ 
Gemini-1.5-pro~\cite{team2023gemini} & 52.70 & 33.20 & 28.33 & 19.20 & 29.19 & 33.36 \\
Cambrian-1-34B~\cite{tong2024cambrian} & 55.00 & 36.00 & 19.48 & 16.07 & 27.06 & 31.64 \\
\bottomrule
\end{tabularx}%
\end{table}

\begin{figure*}[!t]
% \centering
% \footnotesize
\centering\includegraphics[width=1.0\linewidth]{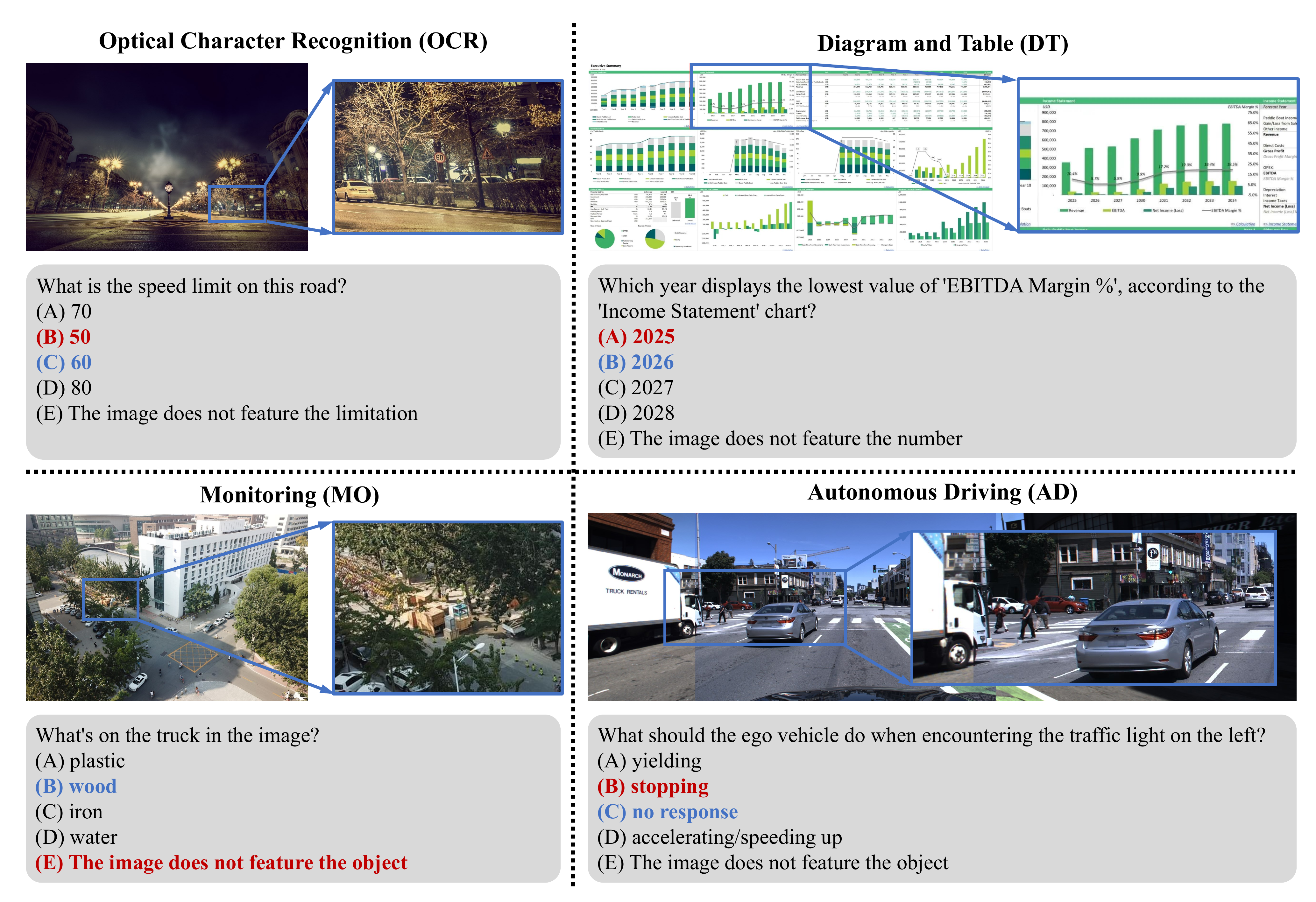}
\vspace{-20pt}
\caption{Representative examples from the MME-Realworld benchmark show InternVL3-78B with our method (answers in \textcolor{RoyalBlue}{blue}) achieving correct deductions where InternVL3-78B without our method (answers in \textcolor{Mahogany}{red}) fails.}
\vspace{-10pt}
\label{fig_results}
\end{figure*}

\renewcommand\arraystretch{0.93}
\begin{table}[h]
\caption{{Experimental results on the perception tasks of the MME-Realworld benchmark.}  OCR, RS, DT, MO, and AD each indicate a specific task: Optical Character Recognition in the Wild, Remote Sensing, Diagram and Table, Monitoring, and  Autonomous Driving, respectively. Avg and Avg-C indicate the weighted and unweighted average accuracy across specific tasks. }\label{tab:mmerw_perception}
\small
\centering
\begin{tabularx}{1.0\textwidth}{lXXXXXXX}
\toprule 
{\textbf{Method}} & \textbf{OCR} & \textbf{RS} & \textbf{DT} & \textbf{MO} & \textbf{AD} & \textbf{Avg} & \textbf{Avg-C} \\  \midrule
% \multicolumn{2}{c}{\textbf{\# QA pairs}}  & 5740 & 3738 & 5433 & 2196 & 3660 & 20767 & 20767 \\ \midrule 
\textbf{InternVL3-78B + Ours} & \textbf{88.15} & \textbf{57.09} & \textbf{83.78} & \textbf{46.68} & \textbf{45.37} & \textbf{69.49} & \textbf{64.21}\\
InternVL3-78B & 85.74 & 56.15 & 78.90 & 45.90 & 40.68 & 66.47 & 61.47 \\
 \midrule 
{InternVL2.5-78B + Ours} & 85.43 & 56.45 & 76.53 & 42.95 & 44.18 & 66.12 & 61.11 \\
InternVL2.5-78B & 81.93 & 55.62 & 70.05 & 41.53 & 40.49 & 62.51 & 57.92 \\
\hline 
Qwen2-VL-7B~\cite{wang2024qwen2} & 81.38 & 44.81 & 70.18 & 37.30 & 34.62 & 58.96 & 53.66 \\
InternVL2-7B~\cite{chen2024internvl} & 73.92 & 39.35 & 62.80 & 53.19 & 35.46 & 55.82 & 52.94 \\ 
Claude 3.5 Sonnet\footnote{\url{https://www.anthropic.com/news/claude-3-5-sonnet}} & 72.47 & 25.74 & 67.44 & 32.19 & 40.77 & 52.90 & 47.72 \\
InternLM-XComposer2.5-7B~\cite{zhang2023internlm} & 69.25 & 36.12 & 63.92 & 39.48 & 33.63 & 52.47 & 48.48 \\
InternVL-20B~\cite{chen2024internvl} & 71.51 & 33.55 & 55.83 & 51.16 & 31.42 & 51.36 & 48.69 \\
Mini-Gemini-34B-HD~\cite{li2024mini} & 69.55 & 40.40 & 44.36 & 39.61 & 32.70 & 48.05 & 45.32 \\
MiniCPM-V2.5-8B~\cite{hu2024minicpm} & 66.79 & 27.69 & 52.81 & 38.70 & 34.15 & 47.37 & 44.03 \\
Cambrian-1-34B~\cite{tong2024cambrian} & 66.45 & 38.63 & 40.44 & 45.98 & 33.61 & 46.68 & 45.02 \\
GPT-4o~\cite{hurst2024gpt} & 77.69 & 28.92 & 46.68 & 33.93 & 22.43 & 46.43 & 41.93 \\
CogVLM2-8B~\cite{hong2024cogvlm2} & 69.97 & 28.76 & 47.51 & 33.74 & 30.22 & 45.84 & 42.04 \\
Cambrian-1-8B~\cite{tong2024cambrian} & 58.68 & 40.05 & 32.73 & 47.68 & 38.52 & 43.82 & 43.53 \\
SliME-8B~\cite{zhang2024beyond} & 53.45 & 42.27 & 29.34 & 40.62 & 33.66 & 40.29 & 39.87 \\
Gemini-1.5-pro~\cite{team2023gemini} & 67.62 & 13.99 & 39.90 & 31.11 & 26.64 & 39.63 & 35.85 \\ 
GPT-4o-mini~\cite{hurst2024gpt} & 62.51 & 6.69 & 44.23 & 26.50 & 24.18 & 37.12 & 32.82 \\
Monkey-7B~\cite{li2023monkey} & 54.63 & 24.99 & 32.51 & 28.01 & 29.67 & 36.30 & 33.96 \\
mPLUG-DocOwl-1.5-7B~\cite{hu2024mplug} & 51.15 & 23.71 & 29.34 & 24.97 & 28.28 & 33.71 & 31.49 \\
DeepSeek-VL-7B~\cite{lu2024deepseek} & 49.55 & 25.49 & 23.38 & 26.97 & 33.39 & 33.14 & 31.76 \\
\bottomrule
\end{tabularx}%
% \vspace{-0.4cm}
\end{table}

\renewcommand\arraystretch{0.93}
\begin{table*}[h]
  \small
  \centering
\caption{Experimental results on the HR-Bench benchmark's 4K and 8K versions across Fine-grained Single-instance Perception (FSP) and Fine-grained Cross-instance Perception (FCP) tasks. "Avg" indicates the weighted average accuracy between the two tasks. } 
\label{table:hr_bench}
  \begin{tabularx}{0.8\textwidth}{@{}lXXXXXX@{}}
  \toprule
  \multirow{2}{*}{Method}     & \multicolumn{3}{c}{{HR-Bench 4K}} & \multicolumn{3}{c}{{HR-Bench 8K}} \\ 
  \cmidrule(lr){2-4} \cmidrule(lr){5-7}
& {FSP}    & {FCP}    & {Avg}    & {FSP}    & {FCP}    & {Avg}    \\ \midrule
  \textbf{InternVL3-78B + Ours} & \textbf{89.3} & \textbf{72.8} & \textbf{81.0} & \textbf{81.0}  & \textbf{67.3}  & \textbf{74.1}  \\
  InternVL3-78B~\cite{zhu2025internvl3}        & 84.3  & 69.3  & 76.8  & 72.8  & 65.5  & 69.1  \\
   \midrule 
  {InternVL2.5-78B + Ours} & {85.3} & {72.0} & {78.6} & {73.8}  & {68.8}  & {71.3}  \\
   InternVL2.5-78B~\cite{chen2024expanding} & 83.3 & 69.8 & 76.5& 71.5 & 65.3 & 68.4 \\
   \midrule 
  InternVL2-llama3-76B~\cite{chen2024internvl} & 82.0 & 60.0& 71.0 & 69.0 & 53.8 & 61.4 \\
  InternVL-1.5-26B~\cite{chen2024far}        & 69.5   & 51.8  & 60.6  & 69.3  & 48.5   & 57.9  \\
  Xcomposer2-4kHD-7B~\cite{dong2024internlm} & 63.8 & 51.8 & 57.8 &  55.3  & 47.3   & 51.3  \\
  LLaVA-1.6-34B~\cite{liu2024llava}       & 55.3  & 50.5   & 52.9  & 44.5   & 50.3  & 47.4  \\
  LLaVA-HR-X-13B~\cite{luo2024feast}      & 61.3  & 46.0     & 53.6  & 49.5   & 44.3  & 46.9  \\
  LLaVA-HR-X-7B~\cite{luo2024feast}       & 57.8  & 46.3  & 52.0     & 42.0     & 41.3  & 41.6  \\
  CogVLM-Chat-17B~\cite{wang2024cogvlm}         & 49.5   & 41.5   & 45.5   & 42.5   & 39.8  & 41.1  \\
  LLaVA-1.6-7B~\cite{liu2024llava}        & 49.0     & 46.8   & 47.9   & 37.2   & 44.2   & 40.8   \\
  Phi3-Vision-4.2B~\cite{abdin2024phi}         & 54.3  & 42.0     & 48.1  & 43.3  & 37.8  & 40.5   \\
  Yi-VL-34B~\cite{young2024yi}           & 46.0     & 42.8  & 44.4  & 39.5   & 38.5   & 39.0     \\
  Yi-VL-6B~\cite{young2024yi}            & 42.8  & 42.5   & 42.6  & 38.5   & 39.3  & 38.9  \\
  LLaVA-1.6-13B~\cite{liu2024llava}       & 49.8  & 41.3  & 45.5   & 38.0     & 38.3  & 38.1  \\
  InternLM-Xcomposer2-7B~\cite{internlmxcomposer2} & 45.5   & 46.5   & 46.0     & 36.0     & 39.8  & 37.9  \\
  LLaVA-v1.5-13B~\cite{liu2024improved}      & 45.2   & 41.3   & 43.3   & 37.5   & 38.0     & 37.8   \\
  SEAL-7B~\cite{wu2024v}                &    47.0    &     29.3   &   38.1 &   42.5     &   28.8     &   35.6       \\
  InternLM-Xcomposer-7B~\cite{zhang2023internlm}           & 37.3   & 41.3   & 39.3   & 34.5   & 35.8   & 35.1   \\
  MPLUG2-7B~\cite{Xu2023mPLUG2AM}              & 38.0     & 35.8  & 36.9  & 33.8  & 33.8  & 33.8  \\
  Deepseek-VL-7B~\cite{lu2024deepseek}         & 36.8  & 34.3  & 35.5   & 33.8  & 33.0     & 33.4  \\
  LLaVA-v1.5-7B~\cite{liu2024improved}       & 38.5   & 33.8   & 36.1   & 33.0     & 31.3   & 32.1   \\
\bottomrule
\end{tabularx}
\end{table*}

\subsection{Evaluation on MME-Realworld Benchmark}
The MME-Realworld benchmark~\cite{mme_realworld_iclr2025} serves as a comprehensive testbed, evaluating MLLMs on a wide array of real-world scenarios. It is divided into perception and reasoning tasks, the reasoning tasks span four challenging tasks: Optical Character Recognition in the Wild (OCR), Diagram and Table Understanding (DT), Monitoring (MO), and Autonomous Driving (AD). These tasks often involve images with complex scenes and require a nuanced understanding of fine-grained details, making it a suitable benchmark to assess the generalizability and detailed comprehension abilities enhanced by Zoom-Refine.

% \subsubsection{Reasoning Results}
As shown in Table~\ref{tab:mmerw_reason}, Zoom-Refine significantly bolsters the reasoning capabilities of MLLMs on the MME-Realworld benchmark. When applied to InternVL3-78B, Zoom-Refine elevates the weighted average accuracy (Avg) by a substantial $3.60\%$, and the unweighted average accuracy(Avg-C) by an even more impressive $4.39\%$. Similar substantial gains are observed with InternVL2.5-78B, with improvements of $4.37\%$ and $5.07\%$ for Avg and Avg-C respectively.
This demonstrates that Zoom-Refine's core mechanism of providing more granular visual information directly translates to an enhanced ability to perform complex logical deductions. For instance, the marked improvement in Diagram and Table (DT) reasoning (e.g., +$9.60\%$ for InternVL3-78B) suggests that the refined visual input allows the model to better understand structural relationships and extract precise data points crucial for analytical reasoning. Similarly, improvements in other tasks indicate a better grasp of spatial relationships and object interactions, which are fundamental to reasoning in dynamic scenes. The uplift across diverse reasoning subtasks underscores that a more faithful and detailed visual understanding, as facilitated by Zoom-Refine, is a critical precursor to robust multimodal reasoning.

Figure~\ref{fig_results} presents qualitative examples from MME-Realworld, illustrating Zoom-Refine's practical impact and underlying mechanisms. For example, in the OCR and DT scenarios, due to information loss from downsampling,leading to misinterpretations of the speed limit and the lowest 'EBITDA Margin \%' year. Our method leverages the MLLM's inherent localization capabilities to identify and zoom into the speed limit sign and the 'Income Statement' chart. By supplementing the initial overview with these high-resolution local crops, the MLLM integrates these fine-grained visual data with the preserved global context, enabling it to refine and correct its preliminary answer without sacrificing broader scene understanding.
In the MO scenario, where the truck is not immediately visible, the MLLM initially hypothesizes its presence in an obscured area. Zoom-Refine facilitates a targeted inspection of such less-clear regions, successfully identifying both the truck and its cargo, thereby overcoming challenges posed by partial occlusion or subtle visual cues.
Similarly, for the AD task, Zoom-Refine's closer examination reveals a critical detail, the green traffic light, that was overlooked during the initial global assessment. This new local evidence, when integrated with the existing understanding of pedestrian presence, 
leading to a refined and more contextually appropriate decision.
In conclusion, these cases demonstrate how the supplementation of localized, high-resolution visual information empowers the MLLM to reflect upon and refine its initial judgments, fostering more accurate and robust reasoning.

% \subsubsection{Perception Results}
Table~\ref{tab:mmerw_perception} presents the results on the perception tasks of MME-Realworld. Our method consistently improves the foundational perception capabilities of the base models. With InternVL3-78B, Zoom-Refine achieves improvements of 3.02\% and 2.74\% for Avg and Avg-C, respectively. Similar positive trends are evident for InternVL2.5-78B, which sees its Avg score rise by 3.61\% and Avg-C by 3.19\%.
These results highlight Zoom-Refine's efficacy in enhancing the fine-grained visual perception of MLLMs. The improvements are not confined to a single task but are observed across various subtasks. Even in OCR, where baseline models are already strong, our method provides further refinement, indicating its general applicability in extracting precise visual information. This enhanced perceptual acuity is crucial, as accurate identification and localization of visual elements form the bedrock for all subsequent cognitive tasks. By enabling models to "see" more clearly and comprehensively, Zoom-Refine pushes SOTA models like InternVL3-78B to new heights in perception across all evaluated subtasks.

\subsection{Evaluation on HR-Bench Benchmark}
To further validate the robustness and general applicability of Zoom-Refine on images with very high resolutions, we also evaluate our method on the HR-Bench benchmark~\cite{wang2025divide}. While MME-Realworld already presents significant high-resolution challenges, HR-Bench is specifically curated with 4K and 8K resolution images and tasks—Fine-grained Single-instance Perception (FSP) and Fine-grained Cross-instance Perception (FCP)—that rigorously probe an MLLM's ability to discern and relate minute details within these expansive visual fields. This provides a complementary evaluation focused on extreme resolutions. Results are presented in Table~\ref{table:hr_bench}.

On HR-Bench 8K, InternVL3-78B enhanced with Zoom-Refine demonstrates a marked increase in average accuracy, climbing by $5.0\%$ from its baseline. Similarly, InternVL2.5-78B sees $2.9\%$ gain. On the 4K tasks, these models achieved respective gains of $4.2\%$ and $2.1\%$ . These results compellingly demonstrate Zoom-Refine's proficiency in handling the unique challenges posed by extremely high-resolution images, where conventional downsampling or naive patch strategies often lead to critical information loss. The consistent uplift across both FSP, which focuses on discerning attributes of individual small objects, and FCP, which requires comparing details across multiple instances or regions, underscores the comprehensive nature of the improvements. Zoom-Refine empowers MLLMs to effectively navigate and interpret many visual data in such images, leading to more accurate perception of minute details and their contextual relationships.

\renewcommand\arraystretch{1.0}
\begin{table}[ht]
\caption{{Comparison of our method with other training-free methods for boosting high-resolution multi-modal understanding.} Experiments are conducted on the reasoning tasks of the MME-Realworld Benchmark using InternVL3-78B as the base model.The 'Time' column indicates the total inference duration for each method to process the benchmark under identical experimental conditions. }\label{tab:comp_sota}
\small
\centering
\begin{tabularx}{1.0\textwidth}{lXXXXXXX}
\toprule 
\textbf{Method} & \textbf{OCR} & \textbf{DT} & \textbf{MO} & \textbf{AD} & \textbf{Avg} & \textbf{Avg-C}  & \textbf{Time}\\ 
\midrule
% \multicolumn{2}{c}{\textbf{\# QA pairs}} & 500 & 500 & 498 & 1344 & 2842 & 2842 \\ \midrule 
\textbf{InternVL3-78B + Ours} & \textbf{75.40} & \textbf{71.80} & \textbf{47.39} & \textbf{40.22} &\textbf{53.22} & \textbf{58.70}  & 447min \\
InternVL3-78B + $\mathrm{DC}^2$~\cite{wang2025divide} & 74.80 & 59.80 & 45.78 & 39.43 & 50.35 & 54.95 &3790min \\
InternVL3-78B + ZoomEye~\cite{shen2024zoomeye} & 74.60 & 61.20 & 46.69 & 38.54 & 50.30 & 55.26 &3572min \\
InternVL3-78B + TextCoT~\cite{luan2024textcot} & 60.20 & 28.80 & 24.70 & 32.59 & 35.40 & 36.57 & 414min \\
InternVL3-78B & 72.80 & 62.20 & 43.78 & 38.47 & 49.62 & 54.31 &{128min} \\
\bottomrule
\end{tabularx}%
\end{table}

\subsection{Comparison with Training-free Methods}
We compare Zoom-Refine with other training-free methods designed to enhance high-resolution image understanding in MLLMs. The comparative evaluation was conducted on the reasoning tasks of the MME-Realworld Benchmark, with InternVL3-78B serving as the consistent base model for all methods. The results are comprehensively presented in Table~\ref{tab:comp_sota}.

Zoom-Refine achieves an average accuracy of $53.22\%$.When compared to other contemporary training-free approaches, it surpasses DC$^2$~\cite{wang2025divide} by $2.87\%$ and ZoomEye~\cite{shen2024zoomeye} by $2.92\%$ percentage points in terms of average accuracy. Indeed, while the architectural designs of DC$^2$ and ZoomEye allow them to effectively supplement details for fine-grained perception, leading to strong performance in OCR, MO, and AD tasks, a contrasting weakness is observed in their handling of the DT task. We observe that DC$^2$ and ZoomEye usually emphasizing localized object regions, may yield incomplete visual context for DT tasks requiring holistic table or chart analysis, thereby limiting their efficacy or even misguiding the model. TextCoT's approach of augmenting cropped images with textual information appears less effective on MME-Realworld. While its crops might be locally relevant, the frequent loss of crucial global context in this complex dataset  curtails its overall performance.
These comparisons underscore Zoom-Refine maintain a highly favorable balance between performance gains and computational cost. Its ability to deliver excellent accuracy with substantially reduced inference time compared to other multi-step methods highlights its practical applicability and efficiency.

\renewcommand\arraystretch{0.93}
\begin{table}[ht]
\caption{{Performance of Zoom-Refine with different sizes of language models.} Experiments are conducted on the reasoning tasks of the MME-Realworld Benchmark using InternVL3-78B as the base model.}\label{tab:diff_size}
\small
\centering
\begin{tabularx}{1.0\textwidth}{lXXXXXXX}
\toprule 
\textbf{Method} & \textbf{OCR} & \textbf{DT} & \textbf{MO} & \textbf{AD} & \textbf{Avg} & \textbf{Avg-C} \\ 
\midrule
% \multicolumn{2}{c}{\textbf{\# QA pairs}} & 500 & 500 & 498 & 1344 & 2842 & 2842 \\ \midrule 
\textbf{InternVL3-78B + Ours} & \textbf{75.40} & \textbf{71.80} & \textbf{47.39} & \textbf{40.22} &\textbf{53.22} & \textbf{58.70} \\
InternVL3-78B & 72.80 & 62.20 & 43.78 & 38.47 & 49.62 & 54.31  \\
 \midrule
InternVL3-14B + Ours & 68.00 & 61.20 & 46.39 & 33.85 & 46.76 & 52.21   \\
InternVL3-14B & 66.60 & 59.80 & 42.77 & 30.88 & 44.34 & 50.01   \\
\midrule
InternVL3-8B + Ours & 67.40 & 45.80 & 43.78 & 37.95 & 45.53 & 48.73   \\
InternVL3-8B & 64.20 & 45.80 & 36.75 & 35.54 & 42.82 & 46.34   \\
\midrule
InternVL3-2B + Ours & 61.00 & 40.40 & 35.94 & 34.15 & 40.29 & 42.87  \\
InternVL3-2B & 59.40 & 41.00 & 22.89 & 32.81 & 37.19 & 39.02  \\
\bottomrule
\end{tabularx}%
\end{table}

\subsection{Analysis of Model Scales}
To assess Zoom-Refine's scalability, we applied it to InternVL3 models of different sizes on the MME-Realworld reasoning tasks, with detailed improvements reported in Table~\ref{tab:diff_size}.
Zoom-Refine significantly delivers performance enhancements across the entire spectrum of model sizes. Notably, uplifts are also registered for its smaller models:  for $+2.42\%$ the 14B model, $+2.71\%$ for the 8B model, and an impressive $+3.10\%$ for the compact 2B model.

It demonstrates that Zoom-Refine's benefits are not solely due to the large size of the MLLMs but provide a more fundamental enhancement to how visual information is processed and fed into the language model. The substantial relative gains for both large and smaller models indicate that the refined visual input from Zoom-Refine allows MLLMs of different sizes to better leverage their existing reasoning abilities, enabling them to achieve performance levels and exhibit capabilities more comparable to those of larger baseline models For smaller models, which might struggle more with complex visual features, the clearer, more detailed input provided by Zoom-Refine can help them to overcome some of their inherent limitations in visual understanding. (e.g., the 2B model's performance on the MO task saw a significant increase by $13.05\%$)

\section{Limitations and Discussion}
\label{sec:limitations}

Although Zoom-Refine demonstrates significant gains through a single application of \textit{Localized Zoom} and \textit{Self-Refinement}, this work does not investigate iterative versions of these processes. An MLLM could potentially benefit from sequential zoom-ins on progressively finer details or different areas of interest, or from multiple rounds of self-refinement, more closely emulating the multi-step, dynamic nature of human visual intelligence. Such iterative approaches might yield further improvements, especially for highly complex scenes or tasks requiring nuanced multi-stage reasoning.

\section{Conclusion}
\label{sec:conclusion}
This paper presents a training-free method named Zoom-Refine that enhances MLLM performance on high-resolution images by mimicking human-like visual processing. Zoom-Refine combines MLLM-guided \textit{Localized Zoom} to focus on relevant details with explicit \textit{Self-Refinement}, where the model critically re-evaluates its initial assessment using the high-resolution crop as evidence against the broader context. This structured self-correction differentiates it from methods merely enriching forward reasoning. Experiments on challenging benchmarks confirmed significant accuracy gains without needing model retraining. While acknowledging limitations like localization dependency and increased latency, Zoom-Refine offers a practical approach to boost detail perception in existing MLLMs, paving the way for more capable visual understanding systems.

\bibliographystyle{plain}
\bibliography{reference}

\end{document}